\documentclass[10pt]{article} 
\usepackage[accepted]{rlc}

\usepackage{amssymb}            
\usepackage{mathtools}          
\usepackage{mathrsfs}           
\usepackage{graphicx}           
\usepackage{subcaption}         
\usepackage[space]{grffile}     
\usepackage{url}                
\usepackage{xcolor}
\usepackage[many]{tcolorbox}
\usepackage{framed}
\usepackage{tabstackengine}
\usepackage{tabularx}
\usepackage{placeins}
\usepackage{wrapfig}

\definecolor{uoftblue}{RGB}{30, 55, 101} 

\newtheorem{theorem}{Theorem}[section]
\usepackage{thmtools, thm-restate}
\newtheorem{definition}[theorem]{Definition}

\title{Can we hop in general? A discussion of benchmark selection and design using the Hopper environment}

\author{Claas Voelcker\\
    cvoelcker@cs.toronto.edu\\
    University of Toronto\\
    Vector Institute, Toronto
    \And
    Marcel Hussing \\
    mhussing@seas.upenn.edu\\
    University of Pennsylvania
    \And
    Eric Eaton \\
    eeaton@seas.upenn.edu \\
    University of Pennsylvania}

\begin{document}

\maketitle

\begin{abstract}
Empirical, benchmark-driven testing is a fundamental paradigm in the current RL community. While using off-the-shelf benchmarks in reinforcement learning (RL) research is a common practice, this choice is rarely discussed. Benchmark choices are often done based on intuitive ideas like ``legged robots'' or ``visual observations''. In this paper, we argue that benchmarking in RL needs to be treated as a scientific discipline itself. To illustrate our point, we present a case study on different variants of the Hopper environment to show that the selection of standard benchmarking suites can drastically change how we judge performance of algorithms. The field does not have a cohesive notion of what the different Hopper environments are representative -- they do not even seem to be representative of each other. Our experimental results suggests a larger issue in the deep RL literature: benchmark choices are neither commonly justified, nor does there exist a language that could be used to justify the selection of certain environments. This paper concludes with a discussion of the requirements for proper discussion and evaluations of benchmarks and recommends steps to start a dialogue towards this goal.\end{abstract}

\section{Introduction}
\label{sec:introduction}

When designing a new algorithm in the wide field of reinforcement learning (RL), a seemingly clear and simple question inevitably arises: 
    \emph{How good is it? }

Theoretical research has made great strides in characterizing the complexity and error bounds of many algorithms under specific assumptions on the MDP structure~\citep{kearns2002near, strehl2006pac, jaksch10aregret, farahmand2010error, lattimore2012pacbounds, dann2015sample, osband2016lower, azar2017minimax, zanette2019tighter, jin2020provably, jin2020rewardfree, jin2021bellmaneluder, domingues2021episodic}. However, once function approximation, large state-action spaces, and exploration are introduced, it is often too difficult to obtain rigorous guarantees without further assumptions~\citep{du2019good, kane2022computational, golowich2024exploration}.
In lieu of mathematical yardsticks, the empirical RL community has used benchmarks and competitive testing to obtain performance estimates of proposed algorithms~\citep{bellemare2013arcade, brockman2016gym, tunyasuvunakool2020dm}.
While the statistical validity of empirical comparisons have received attention \citep{henderson2018deep,agarwal2021deep,patterson2023empirical}, another important problem is less discussed: 
    {\emph{Are our benchmarks representative of a wider set of problems of interest?}}

This paper is a play in two parts. In act one, we provide a technical evaluation. We showcase that picking among two different variants of similar benchmarks, we are unable to replicate algorithm evaluation.  
For this, we investigate four algorithms: Soft-actor critic (SAC)~\citep{haarnoja2018sac}, Model-based Policy Optimization (MBPO) \citep{janner2019mbpo},  Aligned Latent Models (ALM) \citep{ghugare2023simplifying}, and Diversity is All You Need (DIAYN) \citep{eysenbach2018diversity}, chosen to account for  the variety of RL paradigms (model-free, model-based, and reward-free). We highlight that derived insights do not generalize across the different implementations for all these algorithms. 
This shows the field does not have a cohesive notion \emph{of what} the different Hopper environments are representative
---they do not even seem to be representative of each other.

In the second act, we form a position statement by reflecting on the outcome of the technical section.
We argue that our experiments necessitate a reorientation of the community on the role and selection of benchmarks.
In this, we aim to widen the frame of reinforcement learning research to include research on the benchmarks themselves.
The intuitive idea of a ``Hopper'' does evidently not capture the relevant aspects and difficulties of the benchmark.
Without evaluating its benchmarks, the RL community is unable to fully quantify whether there is genuine progress towards the goal of general learning agents.
We summarize the overarching claim of these two acts in a single statement:
\begin{tcolorbox}[boxrule=0.2mm,colback=white,colframe=uoftblue,boxsep=0pt,top=3pt,bottom=5pt]
\centering
\emph{Benchmark selection impacts the evaluation of algorithms, but in the past it has largely been neglected as a first-class problem. This necessitates a re-evaluation of how we describe, compare, and design our RL evaluation platforms.}
\end{tcolorbox}

\section{Background}
\label{sec:background}

Before we discuss whether commonly used benchmarks are adequate, we require at least a rough definition of the term ``benchmark''. Although benchmarks play a vital role in the development of machine learning~\citep{imagenet_cvpr09,raji2021ai}, there is little formal study of their design and role, especially in the context of RL.
Thus, we synthesize an informal definition from a dictionary \citep{dictionary} and prior work asking a similar question to ours in image classification~\citep{raji2021ai}.
\begin{definition}[Benchmark]
  A benchmark is a software library or dataset together with community resources and guidelines specifying its standardized usage. Any benchmark comes with pre-defined, rankable performance metrics meant to enable algorithm comparison.
\end{definition}
To be a little more meticulous about this definition, we can try to draw from other fields that have treated benchmarking as an area of study. In performance engineering, \citet{kistowski2015how} define various lower-level characteristics that are helpful for us to understand what a benchmark is. It should be relevant to the behavior of interest, reproducible, fair, verifiable, and easily usable. While all of these play a crucial role also for the RL community, our focus will be on the first point.

To understand relevance, we differentiate benchmarks from domain-specific test environments.
An example of the latter would be a specific robot simulator for  platform for which an engineer requires a control algorithm.
The engineer might use this simulator of a physical platform, carefully test the difference between the simulation and the real environment, and then design and tune an RL algorithm to produce a controller which is executed in the real environment.
A benchmark however is used not to obtain an instantiation of a policy, but rather to test whether a method to obtain policies performs well on the specified metric.
The important differentiating factor is that an RL researcher does not necessarily care about an agent that, for example, plays Atari games well.
Atari games serve as a proxy for an actual problem of interest interest which may be hard to specify. 

Given that the performance on the specific metric is not the main goal, we instead hope that it is \emph{correlated} to a problem \emph{representative} of a wider class of interesting scenarios within which our finding generalize.
This is is rarely explicitly assessed and there are philosophical difficulties in establishing what qualifies as \emph{representative} \citep{chollet2019measure}.
Notably, in fields like image classification and natural language processing, discussing if and how benchmarks actually represent problems of interest is an active subfield \citep{scheuerman2021do}, yet this discussion is mostly absent in the RL literature.
We defer a further discussion of the nuances of this question to the second part of the paper, and simply posit the following, somewhat testable core requirement for now:
\begin{tcolorbox}[boxrule=0.2mm,colback=white,colframe=uoftblue,boxsep=0pt,top=3pt,bottom=5pt]
\centering
If an algorithm performs well on a benchmark environment, \\ it should also do well on other environments representing similar objectives.
\end{tcolorbox}

\section{Empirical case study: The Hopper environment}
\label{sec:case_study}
For continuous control, a standard testing scenario includes simplified locomotion benchmarks. In these, the goal is to make various instantiations of legged robots move forward. There are two commonly used variants of this benchmark, the OpenAI Gym~\citep{brockman2016gym} and the DeepMind Control~\citep{tunyasuvunakool2020dm} suites. Both are built on top of the MuJoCo physics simulator~\citep{todorov2012mujoco} and contain similar robots and tasks, making them well-suited for testing our previous requirement. Our focus will specifically be the Hopper robot, whose goal it is to hop forward without falling over. Paraphrasing our requirement, one might be tempted to say: If it looks like a Hopper and hops like a Hopper, then it better be able to hop everywhere. 

To verify whether algorithms are able to make the Hopper hop, we selected four commonly used algorithms. We chose SAC~\citep{haarnoja2018sac} as one of the most popular model-free RL algorithms, MBPO~\citep{janner2019mbpo} and ALM~\citep{ghugare2023simplifying} as representatives of the model-based paradigm, and DIAYN~\citep{eysenbach2018diversity}, a reward-free RL approach. For details on our implementation choices we refer to Appendix~\ref{app:impl}.
Each algorithm was originally published using the OpenAI gym variant of the benchmark.
We first discuss the major differences between the two implementations and then present a detailed comparison of algorithm performance.

\subsection{Comparing the benchmark specifications}

The Hopper environment is, to the best of our knowledge, first mentioned in \citet{Erez-RSS-11} and it reappears in~\citet{pmlr-v37-schulman15} and \citet{ddpg}.
In the current Gym version~\citep{towers_gymnasium_2023}, the task description in the documentation (May 2024) is \emph{``The Hopper is a two-dimensional one-legged figure that consist of four main body parts - the torso at the top, the thigh in the middle, the leg in the bottom, and a single foot on which the entire body rests. The goal is to make hops that move in the forward (right) direction by applying torques on the three hinges connecting the four body parts.''}~\citep{hopper}
The reward model consists of three terms: a bonus for remaining \emph{healthy}, which is defined with a set of constraints to ensure that the Hopper is roughly upright, a bonus for achieving a forward velocity, and a small regularization cost for large torques.
The episode is also terminated if the Hopper becomes ``unhealthy'', meaning it has toppled over.

The other variant of the Hopper benchmark is a reimplementation of the environment as part of the DeepMind Control Suite of environments~\citep{tunyasuvunakool2020dm} based on \citep{ddpg}.
In this variant, there are three important changes. The Hopper's torso has one additional link, which also adds another controllable joint.
The reward is designed to be strictly contained within the interval $[0,1]$, making comparisons of algorithms simpler as the maximum cumulative return can be computed from the episode length.
Finally, the DMC suite does not use early termination of unhealthy environments, trajectories are truncated after $1000$ steps by default.
These design choices lead to an effectively sparser reward in the DMC Hopper environment, as the Hopper obtains $0$ reward when toppling over, but the episode is not reset.

Both variants of the Hopper environment ostensibly represent a very similar robot and task but we will demonstrate that it does not fulfill our core requirement: neither algorithmic performance evaluation nor qualitative claims about the usefulness of the chosen algorithms generalize between these two environments. 
Note that we do not test whether a policy is transferable between simulators, but whether the same algorithm is able to produce a policy on both environments.

\subsection{Reward-based reinforcement learning}
\label{sec:emp_rb}
\begin{figure}[t]
    \centering
    \includegraphics[width=\textwidth]{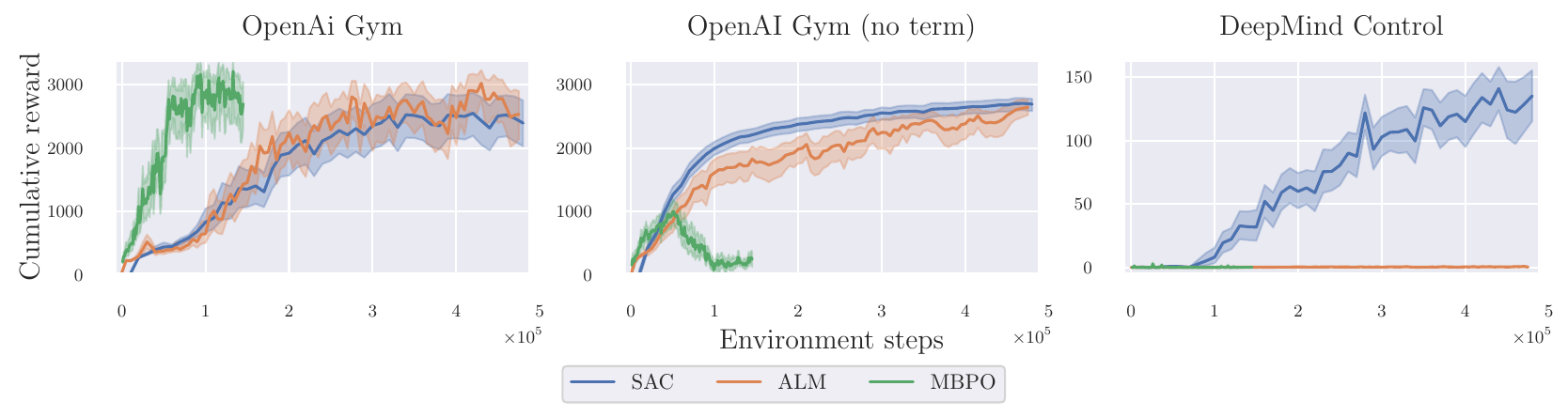}
    \caption{Performance evaluation on the Hopper environment variations. Shaded area denotes a bootstrapped 95\% confidence interval of the mean at 95 across 30 seeds with 5000 resamples.}
    \label{fig:rew_driven_rl}
\end{figure}

We compare the four previously mentioned algorithms in the Hopper environment using three different instantiations: the standard Gym version of the benchmark, a modified Gym version without forced termination (to test whether the gap mostly stems from early termination), and the DM Control version. We report our results in Figure~\ref{fig:rew_driven_rl}. 

SAC is able to solve OpenAI Gym variant of the benchmark, but struggles with the DMC version.
To the best of our knowledge, some recent algorithms achieve at most roughly 500 reward points on this benchmark~\citep{doro2023sampleefficient,hussing2024dissecting,hansen2024tdmpc}.
However, both model-based approaches do not achieve any reward on the DMC variant, even though they achieve on par or better results on the OpenAI Gym version.
This means that depending on which variant of the benchmark is used, neither the absolute performance nor the relative ranking of the algorithms stays intact, a feature that other benchmarks such as ImageNet do possess \citep{recht2019imagenet}.

The hypothesis that this effect can be attributed mostly to early termination cannot be verified.
While MBPO struggles with the absence of termination, both SAC and ALM obtain similar returns to the early terminated variant.
Further investigation on the performance of model-based algorithms in environments without early termination seems pertinent in this light.
For completeness sake, it is important to note that not all model-based approaches fail on the DMC Hopper task. Algorithms such as TD-MPC2 achieve strong performance even without termination \citep{hansen2024tdmpc}.

\subsection{Reward-free reinforcement learning}
\label{sec:emp_rf}

\begin{figure}[b]
    \centering
    \includegraphics[width=\textwidth, trim=0cm 0cm 0cm 3.3cm, clip]{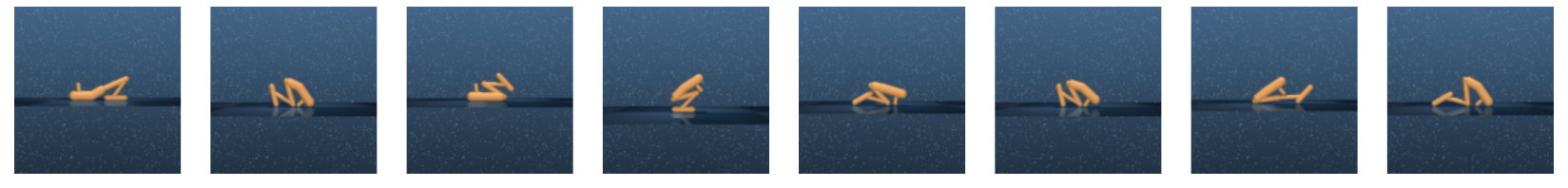}
    \caption{Final states for 8 out of 20 ``skills'' learnt by the DIAYN algorithm on DMC Hopper. In all cases the Hopper immediately moves towards the final configuration without displaying the dynamic skills reported in the original paper.}
    \label{fig:diayn}
\end{figure}

For reward-free RL, we present a qualitative evaluation.
First, we note that the original work presenting the DIAYN algorithm \citep{eysenbach2018diversity} tests their approach on the Hopper variant with early termination.
Their results are presented at this link: \href{https://sites.google.com/view/diayn/hopper}{sites.google.com/view/diayn/hopper}.
In their evaluation, the Hopper performs a variety of different dynamic skills, which can be used to obtain policies for downstream tasks.

We repeated the same experiment now on the DMC variant, without early termination.
We visualize the end states of the trajectories in \autoref{fig:diayn}, with the full trajectories visualized in the appendix.
Without termination, the optimization target of DIAYN does not incentivize learning dynamic skills, but rather distinct static poses, as these are simple to distinguish for the discriminator.
As most common definitions of ``skill'' imply a dynamic motion, not a static contortion, we highlight that the emergence of skills with DIAYN seems to depend on properties of the \emph{testing environment} as much as on the proposed \emph{algorithm}.
Thus, if another variant of the benchmark, had been chosen, the analysis of the algorithm would likely have been different.
This again highlights our core claim.

\section{
How do we describe and evaluate benchmarks?
}

In some sense, the previous section can be narrowly read as a simple comparison of a single environment out of set of two common test benchmarks.
Finding that these two benchmarks behave differently, especially when realizing that not even the observation or action spaces are identical, might not be too surprising. However, dismissing the experiments ignores the role and use of benchmarks in the wider context of the RL literature. The fact that the observation and action spaces are different \emph{should not matter to this degree.} If you agree with our initial requirement, namely that algorithms should work across benchmarks that represent the same agent and have the same objective, we may have convinced you at this point that something is wrong.

If a benchmark is supposed to be representative of a wider set of algorithmic problems, we must begin with investigating the goal of building and using the benchmark.
The question that then demands an answer is: What is the ultimate purpose of reinforcement learning? 
The broad purpose of AI is often stated as the development and understanding of general intelligence, as vague as that might be. 
If this is indeed the intended purpose, the fact that our algorithms performs so differently on benchmarks with arguably similar objectives, is ---at least to us--- somewhat unsettling. 

\subsection{Evaluating the purpose of reinforcement learning}

Reinforcement learning as a field studies methods to compute optimal policies (or related quantities e.g. value functions) in Markov Decision Processes.
The fundamental object here is not derived by a real-world application such as with image classification, but is instead a formal, mathematical object that can be arbitrarily complex.
The generality of the MDP is often seen as a strength of the RL paradigm, as many problems which would intuitively require intelligence can be phrased as a specific variant of the general framework.
However, this generality also causes problems.

If the purpose of our field was simply the mathematical study of MDPs, then empirical research is not needed at all.
However, finding efficient algorithms to solve \emph{any} MDP is likely an intractable proposal.
Instead, we might want efficient solutions to many real world problems by understanding their properties \citep{boss}.
Approaching the overarching purpose from the empirical side as well to establish relevant (sub-)problems on the path towards general intelligence is then indeed important.
We invite the community to approach the resulting question, even though it is difficult: What specific (sub-)problems are important to solve with our algorithms and how do we measure whether our benchmarks properly capture these problems?

We acknowledge the main challenge with our ultimate purpose: Verifying that a benchmark meaningfully represents general intelligence is incredibly difficult. 
Thus, we argue that RL benchmarking needs to be treated as a scientific discipline itself.
This is not a criticism of the hard work that has already been put into developing novel RL benchmark environments.
We value the contributions of such work as establishing new test beds is one way to address this difficult problem. 
To acknowledge a sampling of such efforts, we briefly survey related work on benchmarks in Appendix~\ref{app:related}.
Rather, similar to recent efforts in evaluating Large Language Models, we argue that the RL community needs to start thinking about how to describe its benchmarks and motivate their selection.
In addition to more diverse and interesting benchmarks, we call for expanding the frame of RL research to include assessing the benchmarks themselves.

\subsection{Towards effective benchmarks: Developing a common language}

One of the first steps in this endeavor we propose is the development of a common language for benchmarks.
We need to establish an understanding of important concepts, such as goals, properties, and measurable quantities that are often used without proper specification. In this work, we have already introduced some nomenclature but we would like to emphasize that it might by no means be perfect or the standard from here on out. It is an attempt to start a conversation around a topic we believe to be quite difficult to grasp. We have established the term {\em purpose} for the overarching objective of the community and {\em problems} as concrete challenges on the path towards this goal. 

We use the term {\em property} to describe the intuitive categories that researchers have used to describe and group benchmarks, such as ``legged robot'' or ``visual observation''.
However, our work hopefully convinces you that these are not necessarily useful!
Purely intuitive notions of ``properties'' such as ``Hopper'' cannot be used without establishing and testing whether they represent meaningful groupings with regard to problems of interest. 
We do not know how to do this in all generality but at least we have established that such tests can be attempted.
While some attempts have been made to define properties of MDPs more rigorously \citep{Osband2020Behaviour}, to the best of our knowledge no work has attempted to formalize and study these properties in most established benchmarks.

To develop such properties, we require measurable quantities to group and compare environments, and to test whether they are meaningful for the problem they were defined for.
One positive example that studies properties of benchmarks is \citet{laidlaw2023bridging}. It introduces the ``effective horizon'' as a testable quantity of exploration difficulty in the Atari games.
Another example from the empirical literature is \citet{machado2018revisiting}, who investigate common usage patterns of the ALE suite.
Importantly, these quantities need to be computable across arbitrary benchmarks to allow us to draw conclusion across test-beds.
Moreover, they might be highly entangled with each other and getting to a state of disentanglement will require nuanced treatment and a continually improving research process. For example, when measuring the exploration difficulty in a pixel-based environment, the visual representation problem could be a confounding factor. Extending and verifying these quantities should be a celebrated contribution to the literature at relevant venues. 

Finally, it is necessary for future RL papers to discuss how the chosen benchmarks impact algorithmic design choices.
For example, our experiments show that for DIAYN, early termination is not merely an incidental property of the benchmark that is external to and separate from the algorithm.
Instead, it is important so that the algorithm behaves as intended. By not discussing such properties directly, we structurally blind ourselves to understanding their impact. An outcome of the study of benchmarks should be the ability to inform a conscious choice of the appropriate environments.

\subsection{A peek at other fields of science}

Purpose, goals, properties and measurable quantities have a somewhat circular relationship as one needs to define a purpose to define a goal and corresponding properties and quantities but we require disentangled quantities to reason which goals are achievable.
At this point we would like to draw a parallel to physical sciences. In physics, science often starts out with a first stab to characterize a system. Isaac Newton defined a ``Force'' as a measurable quantity consisting of mass and acceleration. For a long time, this was sufficient to explain various physical behaviors. With the study of more complex systems such as atomic system, these definitions needed to be revisited. We believe that a similar, open-ended research process will be necessary to characterize benchmarks.

Thus, we should allow researchers that attempt to make progress towards this end to publish their work even if it may not capture every single facet of the problem. Similar to how we often take small steps to develop solutions in the algorithm realm, we should allow ourselves to make incremental but consistent progress on the relevance of benchmarks.
Many other fields have worked on the question of how to measure important quantities of interest (take for example the enormous difficulty of establishing a meaningful measure of human intelligence), and how to test whether novel approaches make meaningful progress towards them.
These can serve as inspiration for us, for example reviewing the thorough and rigorous testing standards of engineering fields such as material science.

\section{Conclusion}

We have questioned the motivation for benchmark usage in RL 
and ask the community to consider a meta-level question that will help inform more conscious choices around which benchmarks are relevant for what. More work is needed to properly characterize what our benchmarks \emph{represent}. 

\begin{center}
    \emph{What is the ultimate purpose and what are the underlying goals for the RL community? \\And how do we capture its various aspects in our benchmark environments in a testable manner?}
\end{center}

\clearpage

\subsubsection*{Acknowledgments}
\label{sec:ack}
We thanks the anonymous reviewer for their help, especially with framing the abstract to clarify the positioning of the paper.

This research was partially supported by the Army Research Office under MURI award W911NF20-1-0080, the DARPA Triage Challenge under award HR00112420305, and by the University of Pennsylvania ASSET center. Any opinions, findings, and conclusion or recommendations expressed in this material are those of the authors and do not necessarily reflect the view of DARPA, the Army, or the US government.


\bibliography{main}
\bibliographystyle{rlc}

\newpage

\appendix

\section{Implementation details of algorithms}
\label{app:impl}

\FloatBarrier

\begin{table}[h]
    \centering
    \begin{tabularx}{\textwidth}{>{\hsize=0.14\hsize}X|X}
        Algorithm & Code base\\\hline
        SAC & \url{https://github.com/proceduralia/high_replay_ratio_continuous_control} \\
        MBPO & \url{https://github.com/facebookresearch/mbrl-lib} \\
        ALM & \url{https://github.com/RajGhugare19/alm} \\
        DIAYN & Personal reimplementation based on \url{https://github.com/proceduralia/high_replay_ratio_continuous_control} 
    \end{tabularx}
    \caption{Code sources for tested algorithms}
\label{tab:my_label}
\end{table}

The source code repositories for all algorithms are presented in \autoref{tab:my_label}.
We aimed to make minimal changes to the algorithms and hyperparameters, to stay close to the way that they have been evaluated in previous work.

We used all algorithms with the same hyperparameters across all environments, taken from the original papers proposing each algorithm or the off-the-shelf reference implementations.
We note that all hyperparameters are similar to each other, as all algorithms use SAC as the actor-critic component, and even with some moderate hyperparameter tuning, we were unable to elicit good performance from MBPO and ALM on DMC. 
We also note that while MBPO is seemingly trained for significantly fewer steps, this is in line with the original paper, as MBPO uses a significantly higher update ratio.

All of our chosen algorithms were originally implmented on the OpenAI Gym benchmark variant.
This was a conscious choice, for two reasons. 
First, the OpenAI Gym version is the more popular benchmark alternative, which means more algorithm in general are tested on it.
Second, the existence of early termination requires explicit handling in the code.
To minimize the amount of changes to the code we were required to make, it was useful to chose algorithm implementations that already handle this correctly.
An algorithm like T-MPC2 \citep{hansen2024tdmpc} would have required potentially substantial changes to be useable on OpenAI Gym.

\newpage
\section{DIAYN sequences}

\begin{figure}[h!]
    \centering
    \includegraphics[width=\textwidth]{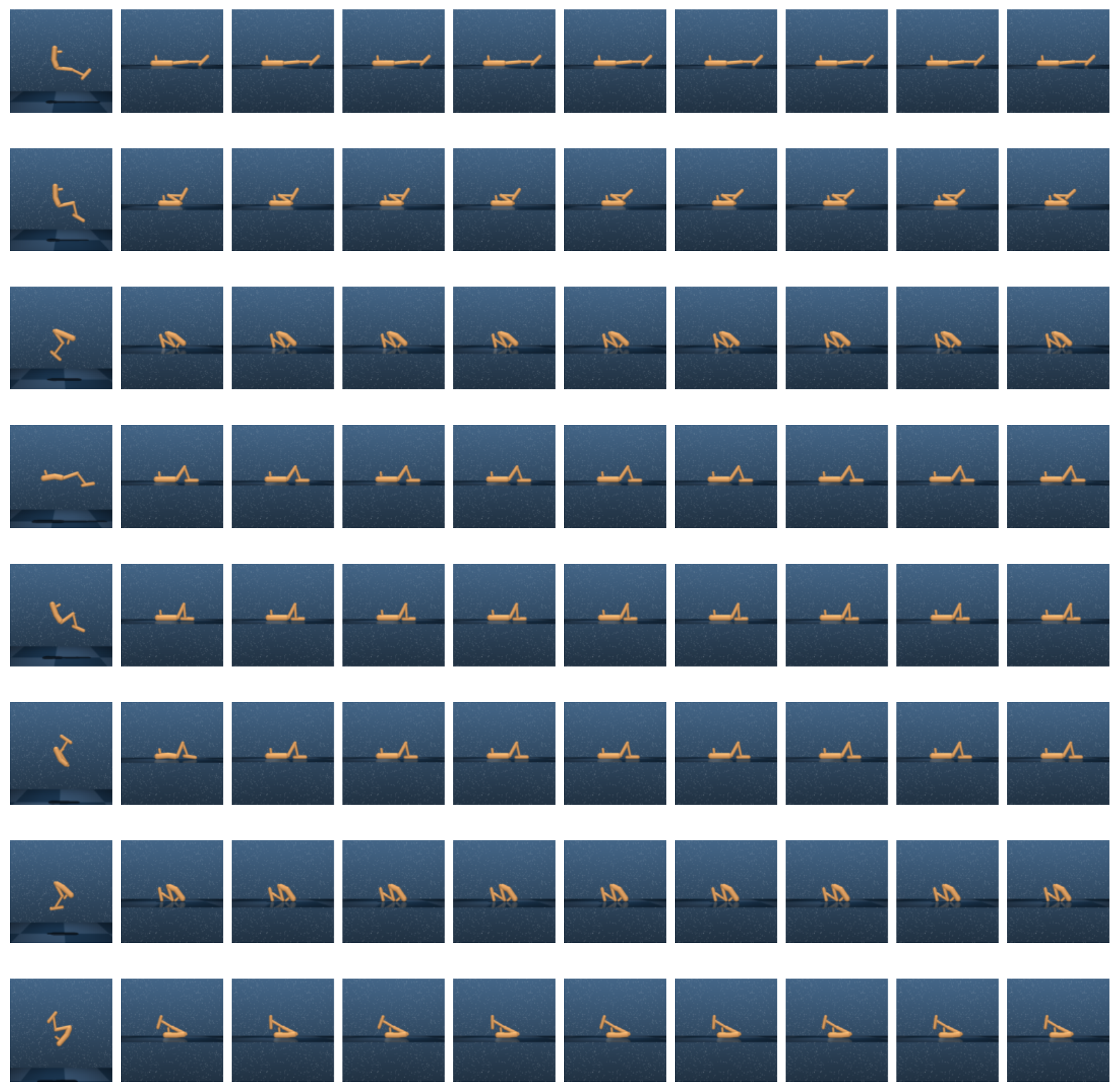}
    \caption{Full visualization of 8 out of 20 skills from one run of DIAYN. Each row is one sequence from the execution of a skill, with each column one frame every 100 environment time steps. For almost all skills we observe that the Hopper maintains a static pose after the first 50-100 time-steps, which is a very different behavior from the one reported in the original paper.\\The poses are mostly distinct, which does optimize the discriminative objective of the algorithm.
    }
    \label{fig:enter-label}
\end{figure}

\section{Additional related work on benchmarks} \label{app:related}

The development of new testbeds has played a significant role in the area of reinforcement learning. The following is a non-exhaustive list of highly valuable benchmark contributions to the field of RL.

A classical application for benchmarking in RL has always been video games due to the planning complexity and our ability to quickly simulate them~\citep{bellemare2013arcade, brockman2016gym, kempka2016vizdoom, vinyals2017starcraft, LanctotEtAl2019OpenSpiel, kurach2020google}.
When moving to continuous control, benchmarks are often inspired by robotics application. This includes for instance the set of DM control tasks used in this work~\citep{tunyasuvunakool2020dm}. Yet, there are many other attempts to capture the difficulties of continuous control in single task learning~\citep{yuke2020robosuite, james2020rlbench}. To build challenges beyond the single-task realm, others have tried to capture modes beyong single task learning. There exist benchmarks for multi-task~\citep{henderson2017multitask}, meta~\citep{yu2019meta}, continual~\citep{wolczyk2021continual, tomilin2023coom, liu2023libero}, compositional~\citep{mendez2022composuite}, skill~\citep{mu2021maniskill} and causal~\citep{ahmed2021causalworld} learning. 
Further, researchers have integrated other challenges posed by different modalities such as text~\citep{cote18textworld, chevalier2019baby, kuettler2020nethack} or visual generalization~\citep{cobbe2019procgen} as well as challenges posed by human-robot collaboration~\citep{puig2024habitat} into the their benchmarks. There are also benchmarks that are more application-based and try to solve automatic algorithm configuration~\citep{eimer2021dac}, combinatorial problems~\citep{bonnet2024jumanji} or financial trading~\citep{liu2021finrl}.
Finally, there are benchmarks which focus on tasks other than reward maximization, such as safe control \citep{Ray2019,9849119}.

\end{document}